\documentclass{article} %
\usepackage{iclr2027_conference,times}

\usepackage{amsmath,amsfonts,bm}

\def\eqref#1{equation~\ref{#1}}

\def\1{\bm{1}}

\DeclareMathAlphabet{\mathsfit}{\encodingdefault}{\sfdefault}{m}{sl}
\SetMathAlphabet{\mathsfit}{bold}{\encodingdefault}{\sfdefault}{bx}{n}

\usepackage{hyperref}
\usepackage{url}
\usepackage{enumitem}
\usepackage{graphicx}

\usepackage{multirow}
\usepackage{booktabs}
\usepackage[table]{xcolor}
\usepackage{fontawesome5}

\title{OmniRay: Camera-Agnostic Monocular 3D Reconstruction with Composite Ray Fields}

\author{Namitha Guruprasad$^{1}$\quad Abhay Yadav$^{1}$\quad Cheng Peng$^{2}$\quad Rama Chellappa$^{1}$\setcounter{footnote}{1}\thanks{Equal advising.}\quad Anand Bhattad$^{1}$\footnotemark[2]\vspace{10pt}\\ 
$^{1}$Johns Hopkins University \qquad $^{2}$University of Virginia
}

\iclrfinalcopy %
\begin{document}

\maketitle
\fancyhead{}
\renewcommand{\headrulewidth}{0pt}
\vspace{-20pt}
\begin{center}
    \href{https://nam1410.github.io/omniray/}{\faCamera\ \textcolor{blue}{\footnotesize{Project Page: https://nam1410.github.io/omniray/}}}
\end{center}

\begin{figure}[h]
\vspace{-10pt}
    \centering
    \includegraphics[width=\linewidth]{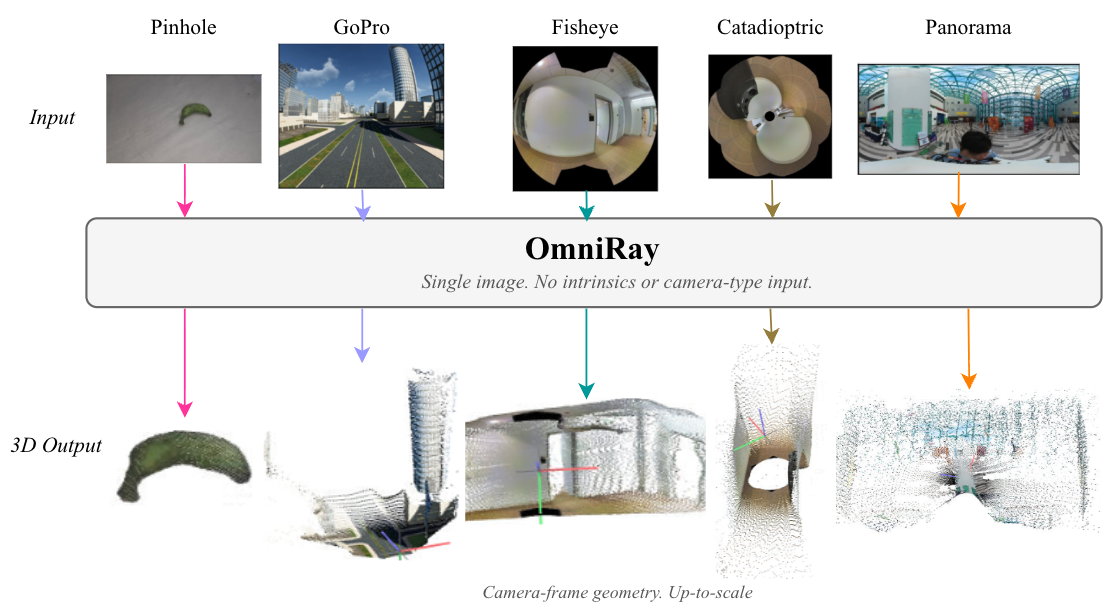}
\vspace{-15pt}
\caption{\textbf{OmniRay reconstructs 3D geometry from a single image across diverse camera geometries.} A single model predicts an up-to-scale, camera-frame point cloud without camera intrinsics or camera-type input. Columns show inputs and reconstructions for pinhole, GoPro, fisheye, catadioptric, and panoramic cameras, respectively. Note that the GoPro and catadioptric camera configurations are withheld from training.}
    \label{fig:teaser}
    
\end{figure}

\begin{abstract}
\sloppy
Monocular models recover 3D geometry from a single image in one pass, but most assume a pinhole camera and produce distorted geometry on the wide-angle imagery produced by fisheye and panoramic lenses. Camera-aware models remove this assumption by predicting a per-pixel ray field. However, they represent different camera geometries using a \textit{single fixed basis} to capture both the overall camera projection and its finer variations. We argue that camera geometry can be viewed as having a hierarchy of complexity for representing rays. We present \textbf{OmniRay}, a feed-forward model whose Ray Module is composed of three components: a monotone radial base that captures radially symmetric structure, a spherical-harmonic residual for smooth deviations from it, and a free-form residual for remaining local variation. These components are composed as successive rotations on the sphere, allowing the representation to accommodate diverse camera geometries. Given a single image, OmniRay predicts a ray direction and an up-to-scale radial distance for every pixel, from which a 3D point cloud is recovered without needing camera information at inference time. 
Extensive experiments across different camera configurations demonstrate that OmniRay establishes a new state-of-the-art in camera-agnostic monocular 3D reconstruction.
\vspace{-5pt}
\end{abstract}

\section{Introduction}
\label{sec:intro}

Monocular 3D reconstruction recovers a 3D point for each pixel in an image. 
We consider central cameras, in which all viewing rays pass through a single optical center, so each 3D point splits into a ray direction and a distance along that ray. The ray direction depends only on the camera geometry, while the distance depends on the scene under observation. Recent feed-forward models perform well on monocular 3D estimation~\citep{yang2024depth, wang2025moge}, but their architectures or training data focus on pinhole cameras. When applied to wide-angle images, these assumptions often lead to distorted reconstructed geometry. Wide-angle cameras are common in robotics~\citep{yogamani2019woodscape}, augmented and virtual reality~\citep{engel2023project}, and $360^\circ$ capture~\citep{liao2022kitti}, making this a practical limitation.
One way to handle these images is to calibrate the camera and rectify them to a pinhole view before reconstruction, but a pinhole cannot represent rays at or beyond $90^\circ$ from its optical axis. As such, undistortion typically leads to cropping away part of the field of view or strongly stretching the image near its periphery, leading to low-confidence predictions from conventional depth estimation or 3D foundation models~\citep{guo2026panovggt}.

Camera-aware models instead account for camera geometry directly~\citep{bogdan2018deepcalib,tirado2025anycalib,piccinelli2025unik3d}. DeepFocal~\citep{workman2015deepfocal} and DeepCalib~\citep{bogdan2018deepcalib} first use convolutional neural networks (CNNs) to estimate focal length and distortion of fixed camera models. Follow-up works generalize camera parameter estimation to arbitrary camera models by estimating ray fields directly. 
In particular, UniK3D~\citep{piccinelli2025unik3d} predicts a per-pixel ray field using a spherical-harmonic (SH) basis and a radial distance along each ray. It can therefore reconstruct pinhole, fisheye, and panoramic images without camera metadata. In this formulation, the same basis must describe both the overall camera projection and its finer variations. We argue that there exists a hierarchy of complexity in camera geometry estimation, which can lead to more efficient and performant design. Many camera projections have a simple structure: for approximately radially symmetric lenses, a smooth radial function is sufficient to describe how viewing directions change across the image. Spherical harmonics can be used to complement radial functions. %

We therefore propose \textbf{OmniRay} -- that takes a single image and predicts a dense, up-to-scale 3D reconstruction without intrinsics or camera-type input (Fig.~\ref{fig:teaser}). A shared image encoder feeds two modules. The Ray Module predicts a monotone radial base and the two residuals, each of which refines the preceding ray field. The Distance Module predicts the radial distance of each pixel, and the corresponding 3D point is obtained by scaling its predicted ray. We train OmniRay on cameras ranging from pinhole to panorama and evaluate it on held-out scenes, datasets, and camera configurations.
An oracle analysis examines which ray fields each representation can express, while benchmark experiments and ablations test the performance of the proposed components.

In summary, our contributions are:
\begin{enumerate}[leftmargin=1.8em, itemsep=0pt]
\vspace{-5pt}
\item A composite ray representation built from a monotone radial base, a spherical-harmonic residual, and a free-form residual.
\item \textbf{OmniRay}, a feed-forward model for single-image 3D reconstruction across central camera geometries, without needing intrinsics or camera-type input.
\item \looseness=-1 An oracle analysis of ray representations, and experiments across seen and held-out scenes, datasets, and camera configurations show that OmniRay improves over the evaluated baselines in ray field estimation and monocular 3D reconstruction.
\end{enumerate}
\vspace{-8pt}

\section{Related Work}
\label{sec:related}

\paragraph{Camera calibration.}
\looseness=-1 Camera calibration determines the mapping from image pixels to rays in the camera frame. Classical methods parameterize this mapping with a low-dimensional model whose parameters are estimated from geometric observations 
~\citep{zhang2000flexible, scaramuzza2006toolbox}.
Different camera geometries are handled by different parametric models: pinhole model with radial and tangential distortion for conventional lenses, \cite{kannala2006generic} model for wide-angle optics, and unified camera models~\citep{geyer2000unifying, mei2007single, khomutenko2015enhanced, usenko2018double} for omnidirectional cameras.
Each of these models fixes a particular functional form, so a lens whose behavior departs from that form introduces residual error. Generic models relax these assumptions with more degrees of freedom and capture complex lens behavior~\citep{schops2020having} at the cost of a higher-dimensional calibration problem.

Learning-based methods offer an alternative to this high-dimensional calibration process by learning camera geometry directly from images without input calibration. DeepCalib~\citep{bogdan2018deepcalib} regresses camera parameters, while Perspective Fields~\citep{jin2023perspective}, GeoCalib~\citep{veicht2024geocalib}, and AnyCalib~\citep{tirado2025anycalib} move toward dense geometric cues: a per-pixel up-vector and latitude field~\citep{jin2023perspective, veicht2024geocalib}, or per-pixel rays~\citep{tirado2025anycalib}, each of which describes the camera more generally than a parameter vector.
Neural Ray Surfaces~\citep{vasiljevic2020neural} take it further by learning a free-form ray surface from video observations and directly representing the pixel-to-ray mapping.
These calibration approaches shift from explicit parameter estimation toward learning the pixel-to-ray mapping, but leave open \textit{how this mapping should be represented to remain both compact and expressive simultaneously}.

\vspace{-8pt}
\paragraph{Monocular 3D.} Recent foundation models have improved depth estimation through large-scale training and distillation~\citep{yang2024depth}, but do not explicitly model the camera geometry. Depth Anything 3~\citep{lin2025depth} instead predicts depth with per-pixel rays by modeling camera geometry alongside scene structure. Camera-aware models specifically account for camera geometry in monocular depth estimation. UniK3D~\citep{piccinelli2025unik3d} represents a per-pixel ray field with spherical-harmonics and predicts radial distance along each ray to handle diverse camera geometries without camera metadata. Depth Any Camera~\citep{guo2025depth} and UniDAC~\citep{ganesan2026unidac} instead transform images into a common equirectangular representation to process different camera types before depth estimation, but require input camera intrinsics for the transformation. CAM3R~\citep{guruprasad2026cam3r} and Wid3R~\citep{jung2026wid3r} extend the spherical-harmonic ray representation to multi-view feed-forward 3D reconstruction. Together, these works highlight the value of explicitly accounting for the pixel-to-ray mapping for camera-agnostic 3D geometry, while leaving open \textit{whether a single ray representation is necessary for all camera geometries}.

\section{Method}
\label{sec:method}
\begin{figure}
    \centering
    \includegraphics[width=\linewidth]{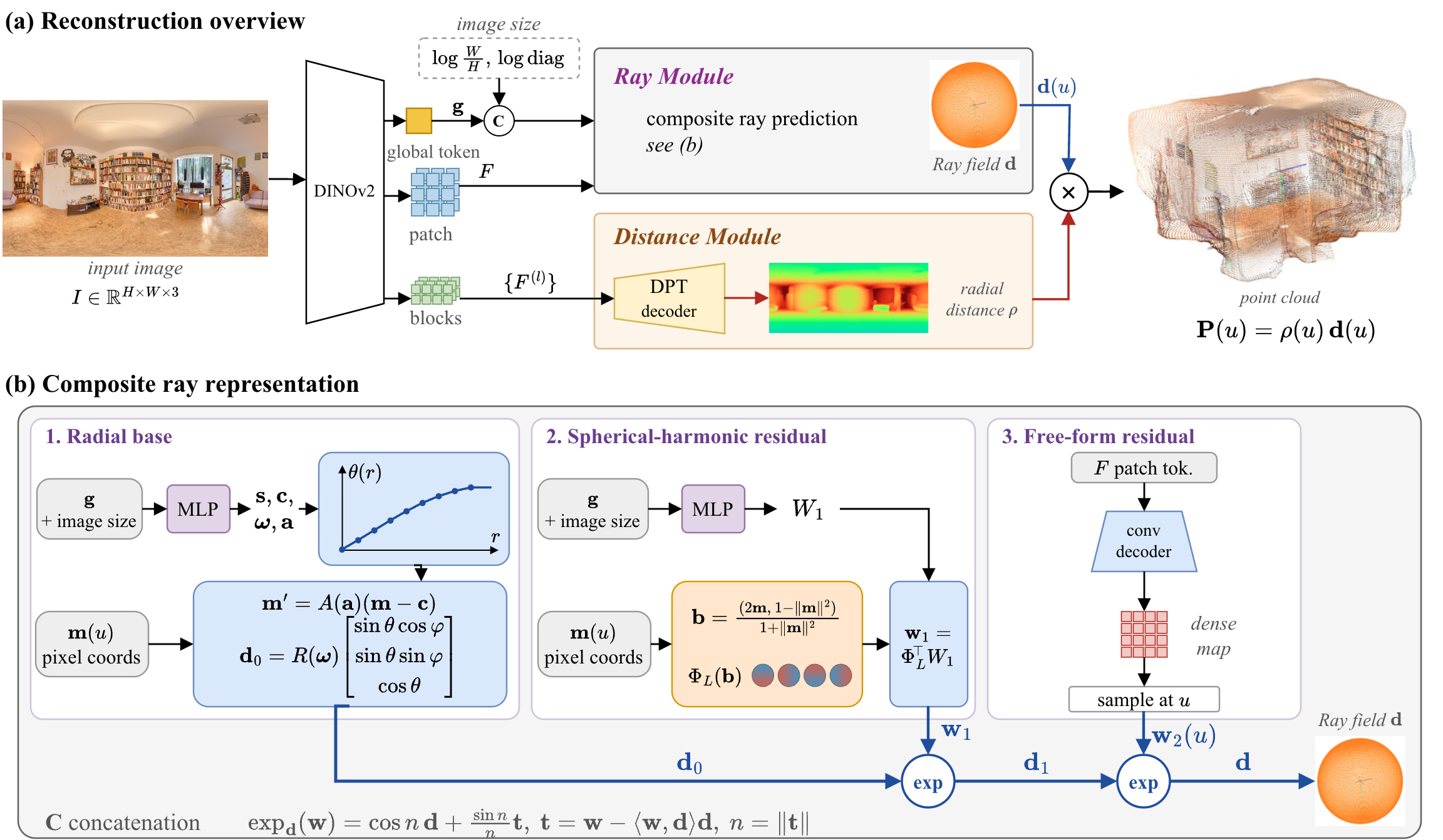}
\vspace{-5pt}
    \caption{\textbf{OmniRay Overview.} Given a single image, a DINOv2 ViT-L/14 encoder produces global and dense features that drive two branches. The \textbf{\textit{Ray Module}} progressively builds a camera-agnostic ray field from a monotone radial base, a spherical-harmonic residual, and a free-form residual. The \textbf{\textit{Distance Module}} predicts a positive radial distance for each pixel. Combining the final ray $\mathbf{d}$ and distance $\rho$ yields the local pointmap $\mathbf{P}=\rho\,\mathbf{d}$, recovering 3D geometry up to scale without camera intrinsics or a camera-type label.}
    \label{fig:arch}
\vspace{-10pt}
\end{figure}

Given a single image $I\in\mathbb{R}^{H\times W\times3}$ from an arbitrary central camera, with no intrinsics, distortion parameters or camera-type label, OmniRay predicts for every pixel $u=(u_x,u_y)$ a unit ray $\mathbf{d}(u)\in S^2$ and a radial distance $\rho(u)>0$. 

A DINOv2 backbone~\citep{oquab2023dinov2} encodes the image, and its features drive two branches (Fig. \ref{fig:arch}): the \textbf{Ray Module} composes $\mathbf{d}$ from radial, spherical-harmonic, and free-form components, while the \textbf{Distance Module} predicts $\rho$. Their combination gives the 3D point in the camera frame:
\begin{equation}
\mathbf{P}(u) = \rho(u)\,\mathbf{d}(u) .
\label{eq:point}
\end{equation}
We predict radial distance $\rho$ rather than perpendicular depth $Z$, related by $\rho=Z/d_z$, since $d_z\to0$ as the ray approaches $90^\circ$ from the optical axis and changes sign beyond it; $\rho$ therefore remains well-defined across wide fields of view.

\paragraph{Notation.}
The DINOv2 backbone produces a dense feature map $F$ and a pooled global token $\mathbf{g}$. We use the normalized image coordinate $\mathbf{m}(u)=(u-c_{\mathrm{img}})/D$, where $c_{\mathrm{img}}$ is the image center and $D$ is the image half-diagonal in pixels. This normalization places the image corners at $\lVert\mathbf{m}\rVert=1$ for any resolution and aspect ratio. Image scale is provided separately to $\mathbf{g}$ through $\log(W/H)$ and the logarithm of the image diagonal. Ground-truth quantities are marked with a star, $\mathbf{d}^\star$ and $\rho^\star$, and $V$ denotes the set of pixels at which they are defined.

\vspace{-5pt}
\subsection{Ray Module}
\label{sec:method:ray}
\looseness=-1 The module comprises three stages, $\mathbf{d}_0 \to \mathbf{d}_1 \to \mathbf{d}$, in which the radial base and the spherical-harmonic residual are predicted from the global token $\mathbf{g}$, and the free-form residual from the dense feature map $F$.

\vspace{-7pt}
\paragraph{Radial base} \looseness=-1 models the approximately radially symmetric structure of conventional central cameras with a monotone mapping from image radius to incidence angle. 

Monotonicity restricts the base to injective radius-to-angle maps, so no two radii in the modeled range receive the same incidence angle, and the profile is evaluated directly rather than inverted numerically as a parametric projection model would require.
A head maps $\mathbf{g}$ to $K+8$ values (Fig. \ref{fig:arch}b): $K$ increments $\mathbf{s}\in\mathbb{R}^{K}$ between fixed knot radii $0=r_0<\dots<r_K$, with a principal-point offset $\mathbf{c}$, an axis-angle rotation $\boldsymbol{\omega}$, and three parameters forming an upper-triangular map $A(\mathbf{a})$ for anisotropic scaling and shear. These give the transformed coordinate $\mathbf{m}'=A(\mathbf{a})(\mathbf{m}-\mathbf{c})$, with radius $r$ and azimuth $\varphi$. The increments are passed through a softplus to ensure positivity, accumulated into knot heights, and linearly interpolated,
\vspace{-2pt}
\begin{equation}
\theta_k=\sum_{j\le k}\operatorname{softplus}(s_j),\quad \theta_0=0,
\qquad
\theta(r)=\operatorname{lerp}\big(\operatorname{clip}(r;r_0,r_K);\ \{r_k\},\ \{\theta_k\}\big),
\label{eq:profile}
\end{equation}
so that $\theta(r)$ is strictly increasing within the knot span and constant outside it. We use $K=16$ knots spanning $r\in[0,1.3]$ in half-diagonal units. The corresponding ray is
\begin{equation}
\vspace{-2pt}
\mathbf{d}_0 =
R(\boldsymbol{\omega})
\begin{bmatrix}
\sin\theta(r)\cos\varphi\\
\sin\theta(r)\sin\varphi\\
\cos\theta(r)
\end{bmatrix},
\label{eq:d0}
\vspace{-2pt}
\end{equation}

where $R(\boldsymbol{\omega})$ is the rotation matrix of $\boldsymbol{\omega}$. The largest angle the base can produce is the sum of its increments, so nothing caps $\theta$ at $90^\circ$ and the base already represents lenses whose rays pass the equator.

\vspace{-8pt}
\paragraph{Spherical-harmonic residual.}

\begin{figure}[t!]
\vspace{-10pt} 
    \centering
    \includegraphics[width=\linewidth]{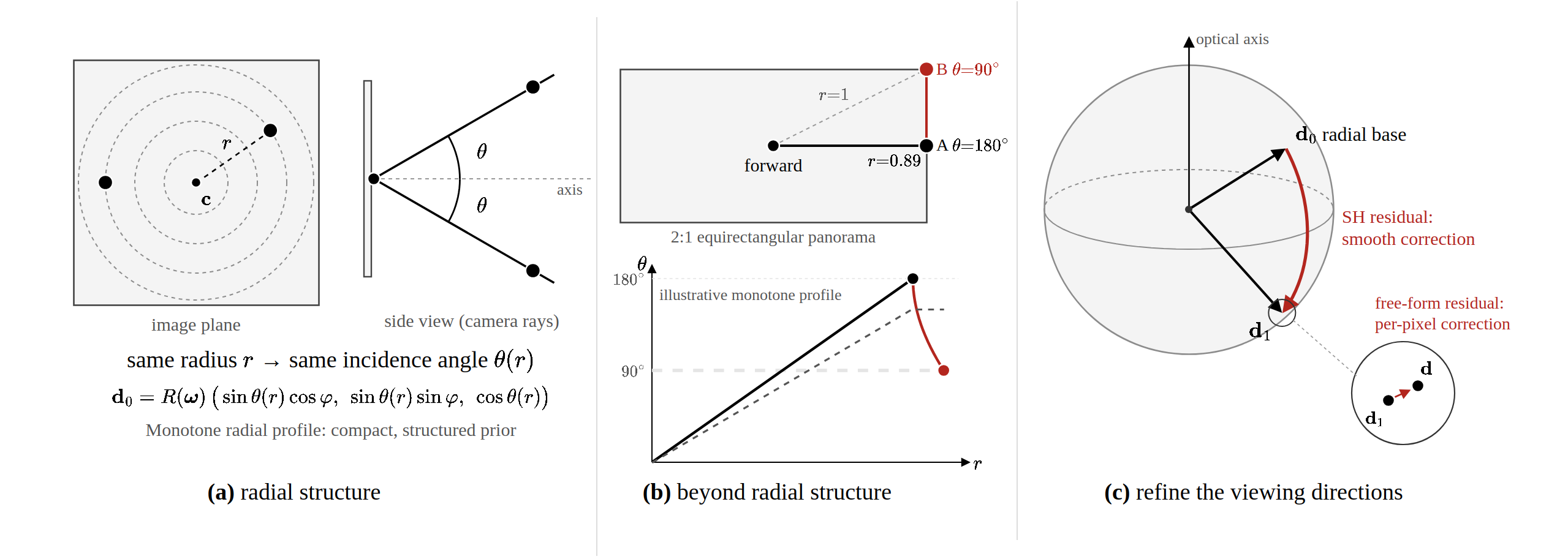}
\vspace{-25pt}
\caption{\textbf{Why combine the radial construction with the residual corrections?}
(a) The radial representation gives the same incidence angle to pixels that lie at the same radius in the base coordinates.
(b) In the centered $2{:}1$ equirectangular projection, point A is closer to the center of the image than point B, yet has a higher incidence angle. A monotone radial function cannot accommodate such ordering under fixed coordinate transformation and orientation.
(c) The spherical-harmonic and free-form residuals improve upon the base direction via iterative correction on the unit sphere.}
 \label{fig:intuit}
\end{figure}

The radial base restricts the incidence angle to a monotone function of image radius, which becomes insufficient when the image covers a large portion of the viewing sphere (Fig. \ref{fig:intuit}b). A spherical-harmonic residual supplies an angular correction to it (Fig. \ref{fig:arch}b). 
A head maps $\mathbf{g}$ to a single coefficient matrix $\mathbf{W}_1\in\mathbb{R}^{(L+1)^2\times3}$ for a real spherical-harmonic basis of maximum degree $L$, shared across all pixels. Thus, the correction field is represented by $3(L+1)^2$ predicted coefficients rather than an independent vector at each pixel. Each pixel evaluates the basis at an address on $S^2$ obtained from its normalized image coordinate through the inverse-stereographic lift,
\vspace{-2pt}
\begin{equation}
\mathbf{b}(\mathbf{m})=\frac{\big(2\mathbf{m},\ 1-\lVert\mathbf{m}\rVert^{2}\big)}{1+\lVert\mathbf{m}\rVert^{2}},
\qquad
\mathbf{w}_1(u)=\mathbf{W}_1^{\!\top}\,\Phi_L\big(\mathbf{b}(\mathbf{m}(u))\big),
\label{eq:sh}
\end{equation}
where $\Phi_L$ stacks the $(L+1)^2$ basis values. We use $L=7$, giving $192$ coefficients per image.

The correction is applied through the exponential map on $S^2$. For any ray $\mathbf{d}$, only the component of $\mathbf{w}$ orthogonal to $\mathbf{d}$ is a tangent direction, so the remaining component is discarded, leaving $\mathbf{t}=\mathbf{w}-\langle\mathbf{w},\mathbf{d}\rangle\mathbf{d}$. Its norm $n=\lVert\mathbf{t}\rVert$ is the rotation angle in radians, and the ray is transported that far along the great circle through $\mathbf{d}$ in the direction of $\mathbf{t}$,

\vspace{-10pt}
\begin{equation}
\exp_{\mathbf{d}}(\mathbf{w})=\cos n\,\mathbf{d}+\frac{\sin n}{n}\,\mathbf{t}.
\label{eq:exp}
\vspace{-1pt}
\end{equation}

\vspace{-8pt}
\paragraph{Free-form residual.}

The spherical-harmonic residual provides a compact global correction but remains limited by the finite-order basis. The free-form residual captures the remaining local variation using a lightweight convolutional decoder on the dense feature map $F$. The resulting correction map is interpolated at each pixel to obtain $\mathbf{w}_2(u)$, so its spatial detail is limited by the resolution of $F$. The two residuals are applied in turn through \eqref{eq:exp}, giving the composite ray field
\begin{equation}
\mathbf{d}_1=\exp_{\mathbf{d}_0}\!\big(\mathbf{w}_1\big),
\qquad
\mathbf{d}=\exp_{\mathbf{d}_1}\!\big(\mathbf{w}_2\big),
\label{eq:compose}
\vspace{1pt}
\end{equation}
with $\mathbf{d}_0$ given by \eqref{eq:d0}, and with $\mathbf{t}_1$ and $\mathbf{t}_2$ the tangential parts of $\mathbf{w}_1$ at $\mathbf{d}_0$ and of $\mathbf{w}_2$ at $\mathbf{d}_1$. Both residual heads are zero-initialized, so the composition starts from the radial base, and the norms $\lVert\mathbf{t}_1\rVert$ and $\lVert\mathbf{t}_2\rVert$ are the 
rotation angles each residual applies, measuring how far a camera departs from the representation beneath it and letting the representation adapt without model selection or routing.

\vspace{-5pt}

\subsection{Distance Module}
\label{sec:method:distance}

The Ray Module determines the direction $\mathbf{d}$ along which each pixel observes the scene, but not how far along that ray the surface lies. The Distance Module predicts this remaining quantity: the up-to-scale radial distance $\rho(u)$ from the optical center. It reuses the same encoder pass, fusing four intermediate blocks with a Dense Prediction Transformer (DPT) decoder~\citep{ranftl2021vision} into a dense prediction sampled at each pixel, where a softplus gives a positive distance $\rho(u)=\operatorname{softplus}(f_\rho(u))>0$. Scaling each predicted ray by its distance as in \eqref{eq:point} produces the camera-frame point cloud $\{\mathbf{P}(u)\}$ in a single forward pass.

\vspace{-5pt}
\subsection{Training objective}
\label{sec:method:loss}

\paragraph{Angular ray loss} supervises the final ray field and the two intermediate fields with the
mean cosine residual $\ell(\mathbf{x})=\frac{1}{|V|}\sum_{i\in V}\big(1-\langle\mathbf{x}_i,\mathbf{d}^\star_i\rangle\big)$,
\begin{equation}
\mathcal{L}_{\mathrm{ray}}=\ell(\mathbf{d})+\gamma_0\,\ell(\mathbf{d}_0)+\gamma_1\,\ell(\mathbf{d}_1)
+\mathcal{L}_{\mathrm{region}} 
\label{eq:lray}
\end{equation}
with stage weights $\gamma_0,\gamma_1>0$. This deep supervision of $\mathbf{d}_0$ and $\mathbf{d}_1$ keeps each stage responsible for the structure it represents, so a correct final ray cannot be reached through an arbitrary base.

$\mathcal{L}_{\mathrm{region}}$ weights the same per-pixel error more heavily where it is largest. 

With $\bar e(S)$ denoting the mean error over a pixel set $S$, it is $\lambda_{\mathrm{c}}\,\bar e(C)+\lambda_{\mathrm{p}}\,\bar e(Q)+\lambda_{\mathrm{t}}\,\bar e(T)$ over an inner region $C$, an outer region $Q$, and tail $T$ of the 5\% of pixels with the largest error, with $\lambda_{\mathrm{p}}$ the largest of the three since error concentrates at the periphery for wide fields of view. The regions are thresholds on the image radius $\lVert\mathbf{m}\rVert$ relative to its maximum, so the split follows the imaged region of a circular fisheye as well as a rectangular frame and does not move with the predicted principal point.

\vspace{-5pt}
\paragraph{point cloud loss} \looseness=-1 supervises rays and distances jointly. Each cloud is normalized by its own mean point norm, $z_{\mathrm{pred}}=\frac{1}{|V|}\sum_{i\in V}\lVert\rho_i\mathbf{d}_i\rVert$ for the prediction and $z_{\mathrm{gt}}=\frac{1}{|V|}\sum_{i\in V}\lVert\rho^\star_i\mathbf{d}^\star_i\rVert$ for the ground truth, which removes the global scale that a single image leaves undetermined,
\begin{equation}
\mathcal{L}_{\mathrm{pc}}=\frac{1}{|V|}\sum_{i\in V}
\Big\lVert\frac{\rho_i\mathbf{d}_i}{z_{\mathrm{pred}}}-\frac{\rho^\star_i\mathbf{d}^\star_i}{z_{\mathrm{gt}}}\Big\rVert_2 .
\label{eq:lpc}
\end{equation}
It is evaluated on $\mathbf{P}=\rho\,\mathbf{d}$ with the ray attached, so a ray error displaces the point even for an exact distance and the gradient reaches both modules.

\vspace{-5pt}

\textbf{Overall objective.} $\mathcal{L}=\lambda_{\mathrm{ray}}\,\mathcal{L}_{\mathrm{ray}}
+\lambda_{\mathrm{pc}}\,\mathcal{L}_{\mathrm{pc}}.$ %
Both modules and the shared encoder are trained end-to-end under this objective. Sec.~\ref{sec:exp:impl} gives the head architectures and the full training schedule.

\section{Experiments}
\label{sec:exp}

\subsection{Oracle test: representation ceilings}
\label{sec:exp:oracle}

Before training anything, we measure what each candidate representation can express at all, by fitting its parameters directly to analytic ground-truth rays; Sec.~\ref{sec:supp:details} gives the fitting protocol. Table \ref{tab:oracle} shows that compact parametric models succeed only on the geometries their assumptions match, and that increasing the number of knots in the monotone radial profile reduces the catadioptric error but leaves the equirectangular error at $22.8^\circ$, the ceiling identified in Sec. \ref{sec:method:ray}. The spherical-harmonic field removes that ceiling but remains two orders of magnitude behind the radial profile on the fisheye. Among the compact representations, only the composition is below $0.25^\circ$ on all three geometries, and since it reduces to the radial base when the residual vanishes, its capacity is bounded below by that of the base.

\begin{table}[t]
\centering
\caption{\textbf{Oracle representation fitting.}
Mean angular error in degrees after fitting each representation
directly to ground-truth rays; lower is better.
\#Params denotes parameters per camera; the SH row adds three rotation parameters to the $3(L+1)^2=192$ coefficients for the base direction, which the residual inherits from the radial base.
Best results among compact representations are \textbf{bold}.
The free field is shown separately as a high-capacity reference.}
\label{tab:oracle}
\vspace{5pt}
\small
\setlength{\tabcolsep}{5pt}
\renewcommand{\arraystretch}{1.12}

\resizebox{0.75\linewidth}{!}{%
\begin{tabular}{@{}lr rrr@{}}
\toprule
& & \multicolumn{3}{c}{\textbf{Angular error} $({}^{\circ})\downarrow$} \\
\cmidrule(l){3-5}
Representation & \#Params
& Fisheye & Equirectangular & Catadioptric \\
\midrule
Pinhole
& 6 & 5.731 & 48.080 & 50.434 \\

Double sphere
& 8 & 0.021 & 22.932 & 41.876 \\

Radial ($K=16$)
& 24 & \textbf{0.002} & 22.802 & 0.104 \\

Radial ($K=32$)
& 40 & \textbf{0.002} & 22.801 & \textbf{0.023} \\

SH ($L=7$)
& 195 & 0.276 & 0.641 & 0.295 \\

\rowcolor{gray!10}
\textbf{Radial$_{16}$ + SH$_7$ residual (ours)}
& 216 & 0.075 & \textbf{0.217} & 0.174 \\

\midrule
Free field
& 76.8k & 0.005 & 0.013 & 0.015 \\
\bottomrule
\end{tabular}%
}
\vspace{-5pt}
\end{table}

\subsection{Datasets and Camera Types}
\label{sec:exp:data}

We train on two panoramic datasets, 2D3DS~\citep{armeni2017joint}, of indoor offices and corridors, and 360Loc~\citep{huang2024360loc}, of larger indoor and outdoor environments, rendering perspective, fisheye, and panoramic views from each panorama; the 2D3DS perspective images are the real captures provided with the dataset. We evaluate on held-out scenes of both, and on two unseen datasets, PanoCity~\citep{guo2026panovggt}, of outdoor street-level captures, and CO3Dv2~\citep{reizenstein2021common}, of object-centric images. Two camera configurations are withheld from training entirely, a GoPro fisheye and a Nayar catadioptric camera~\citep{nayar1997catadioptric,baker1999theory}, both satisfying our central-camera assumption. Sec.~\ref{sec:supp:details} gives the splits and the camera specifications.

\paragraph{Random-lens Augmentation.}
During training we augment each panorama with randomly sampled synthetic central cameras, each defined by a viewing direction, a monotone radial profile drawn from positive spline increments, and an imaged sensor region. This single family covers filled and annular domains, from pinhole to wide-angle lenses. Draws whose profile falls close to a held-out geometry are rejected, so neither withheld camera nor its neighborhood appears in training. Sec.~\ref{sec:supp:details} gives the sampling ranges and the rejection criteria, and Fig.~\ref{fig:randlens} shows sampled examples.

\subsection{Benchmark and Ablation}
\label{sec:exp:bench}

\paragraph{Baselines.} UniK3D~\citep{piccinelli2025unik3d} and Depth Anything~3 (DA3)~\citep{lin2025depth} jointly predict camera geometry and scene structure and are evaluated on all metrics; we fine-tune UniK3D on the 2D3DS and 360Loc panoramas so that it sees the same source imagery we train on. GeoCalib~\citep{veicht2024geocalib} and AnyCalib~\citep{tirado2025anycalib} calibrate the camera without predicting scene geometry. We convert their estimated cameras to dense ray fields for evaluation so all methods are scored identically.

\vspace{-8pt}
\paragraph{Metrics.}
We evaluate the predicted ray fields, radial distance, and 3D point clouds using the following metrics: Ray accuracy is the mean angular error $\arccos\langle\mathbf{d},\mathbf{d}^\star\rangle$ in degrees, and radial distance is evaluated with absolute relative error (AbsRel) and $\delta<1.25$. For point clouds, accuracy and completeness are the median nearest-neighbor distances normalized by the scene's mean point norm, and their mean is the Chamfer distance (CD). Predictions in different depth gauges are aligned per image by $s=\operatorname{median}(\rho^\star)/\operatorname{median}(\rho)$, and perpendicular depth is converted to radial distance by $\rho=Z/d_z$ using the predicted rays. All metrics are computed over pixels with valid ground truth.

\vspace{-8pt}
\paragraph{Results.}

\begin{table*}[t]
\centering
\caption{\textbf{Main benchmark.}
Results on seen cameras, held-out cameras, and held-out domains.
Ang.\ denotes mean ray error in degrees; CD denotes median relative
Chamfer distance, normalized by each scene's mean point norm.
Metrics are macro-averaged over test images, and all methods use
per-image median scale alignment.
Lower is better; best results are \textbf{bold}. AUG denotes random-lens augmentation, and RG denotes ray guidance.
$\dagger$ denotes exclusion from our training data; a dash indicates
that the method does not predict scene geometry.
Full per-camera and per-dataset results and additional metrics
are provided in Sec.~\ref{sec:supp:results}.}
\label{tab:main}
\vspace{5pt}
\small
\setlength{\tabcolsep}{5pt}
\renewcommand{\arraystretch}{1.12}

\resizebox{\linewidth}{!}{%
\begin{tabular}{@{}l *{5}{rr}@{}}
\toprule
& \multicolumn{2}{c}{\textbf{Seen cameras}}
& \multicolumn{4}{c}{\textbf{Held-out cameras}}
& \multicolumn{4}{c}{\textbf{Held-out domains}} \\
\cmidrule(lr){2-3}
\cmidrule(lr){4-7}
\cmidrule(l){8-11}
& \multicolumn{2}{c}{Aggregate}
& \multicolumn{2}{c}{Catadioptric$^\dagger$}
& \multicolumn{2}{c}{GoPro$^\dagger$}
& \multicolumn{2}{c}{PanoCity$^\dagger$}
& \multicolumn{2}{c}{CO3Dv2$^\dagger$} \\
\cmidrule(lr){2-3}
\cmidrule(lr){4-5}
\cmidrule(lr){6-7}
\cmidrule(lr){8-9}
\cmidrule(l){10-11}
Method
& Ang.$\,({}^{\circ})\downarrow$ & CD$\downarrow$
& Ang.$\,({}^{\circ})\downarrow$ & CD$\downarrow$
& Ang.$\,({}^{\circ})\downarrow$ & CD$\downarrow$
& Ang.$\,({}^{\circ})\downarrow$ & CD$\downarrow$
& Ang.$\,({}^{\circ})\downarrow$ & CD$\downarrow$ \\
\midrule
GeoCalib \citep{veicht2024geocalib}
& 30.997 & --
& 119.662 & --
& 14.126 & --
& 47.132 & --
& 3.713 & -- \\

AnyCalib \citep{tirado2025anycalib}
& 28.655 & --
& 107.129 & --
& 10.031 & --
& 42.335 & --
& 2.711 & -- \\

DA3 \citep{lin2025depth}
& 33.028 & 0.1999
& 121.566 & 1.0412
& 16.637 & 0.1097
& 50.533 & 0.3677
& \textbf{2.041} & 0.0288 \\

UniK3D \citep{piccinelli2025unik3d}
& 6.336 & 0.0760
& 75.702 & 0.5948
& 6.342 & 0.0454
& 21.220 & 0.2120
& 2.676 & \textbf{0.0277} \\

\midrule
OmniRay w/o AUG
& 1.385 & 0.0592
& 30.867 & 0.3440
& 7.965 & 0.0641
& 6.352 & 0.0984
& 2.825 & 0.0330 \\

OmniRay + RG
& 0.356 & 0.0205
& \textbf{1.118} & \textbf{0.0113}
& \textbf{5.180} & \textbf{0.0375}
& 1.592 & 0.0277
& 2.935 & 0.0411 \\

\rowcolor{gray!10}
\textbf{OmniRay}
& \textbf{0.140} & \textbf{0.0179}
& 1.120 & 0.0131
& 5.533 & 0.0392
& \textbf{1.350} & \textbf{0.0240}
& 2.967 & 0.0380 \\
\bottomrule
\end{tabular}}
\vspace{-5pt}
\end{table*}

Table \ref{tab:main} summarizes performance across camera configurations and domains. On the seen camera types, OmniRay achieves a mean ray error of $0.14^\circ$ and relative Chamfer distance of $0.0179$, compared with $6.34^\circ$ and $0.0760$ for UniK3D and $33.03^\circ$ and $0.1999$ for Depth Anything~3. On the held-out configurations, for the Nayar catadioptric camera, all baselines exceed $75^\circ$ while ours achieves $1.12^\circ$ and $0.0131$ CD; on GoPro, ours achieves $5.53^\circ$ versus $6.34^\circ$ for UniK3D. On PanoCity, ours achieves $1.35^\circ$ ray error versus $21.22^\circ$ and $50.53^\circ$ for UniK3D and DA3, respectively. On CO3Dv2, our $2.97^\circ$ ray error remains within a degree of both baselines.

\begin{figure}[ht!]
    \centering
    \includegraphics[width=\linewidth]{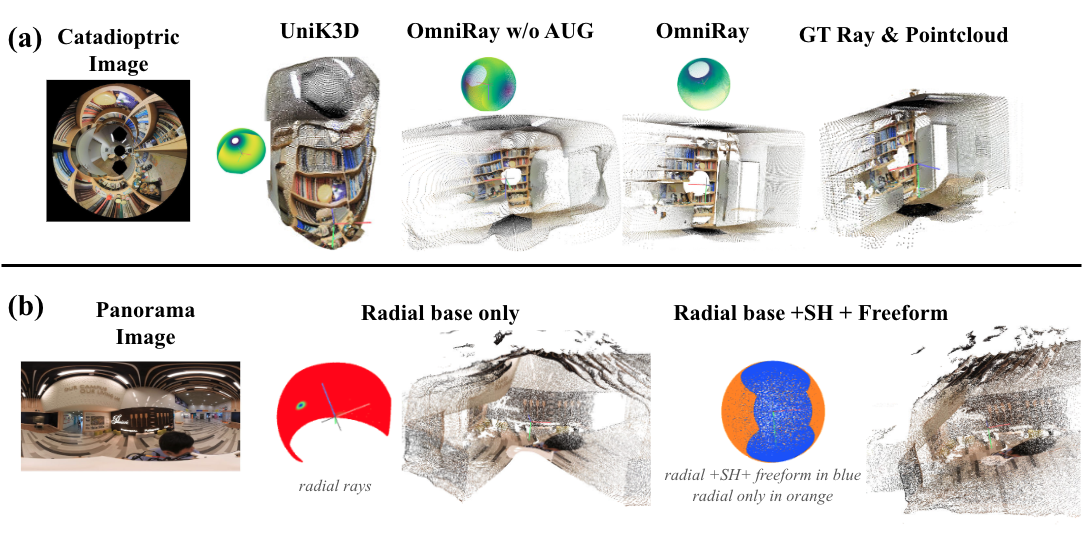}
\vspace{-25pt}
    \caption{\textbf{Ablation.} \textbf{(a)} Held-out catadioptric camera: predicted ray field and point cloud for UniK3D, OmniRay trained without random-lens augmentation, and the full model, against the ground truth. \textbf{(b)} Panoramic image: the radial base alone covers only part of the sphere (orange) and leaves the reconstruction torn, while the spherical-harmonic and free-form residuals complete the field (blue) and close the geometry.}
 \label{fig:ablat}
\end{figure}

The \textbf{w/o AUG} row isolates the random-lens sampling of Sec.~\ref{sec:exp:data}. Without it the seen cameras remain strong at $1.39^\circ$, but the held-out catadioptric and GoPro errors rise to $30.87^\circ$ and $7.97^\circ$; restoring it recovers $1.12^\circ$ and $5.53^\circ$ and improves the seen cameras to $0.14^\circ$. Fig.~\ref{fig:ablat}a shows the same effect on a catadioptric example, where the unaugmented model exhibits larger ray-field and geometric errors. These results indicate that random-lens sampling is important for extending the composite representation beyond the camera configurations encountered during training.
Adding ray guidance to each DPT scale following~\citet{lin2025prompting} (\textbf{+ RG}) does not change the metrics consistently. The injection is zero-initialized, so the guided head could exploit the prompt if it were beneficial, and we therefore omit it from the final model.

\begin{figure}[t!]
    \centering
    \includegraphics[width=\linewidth]{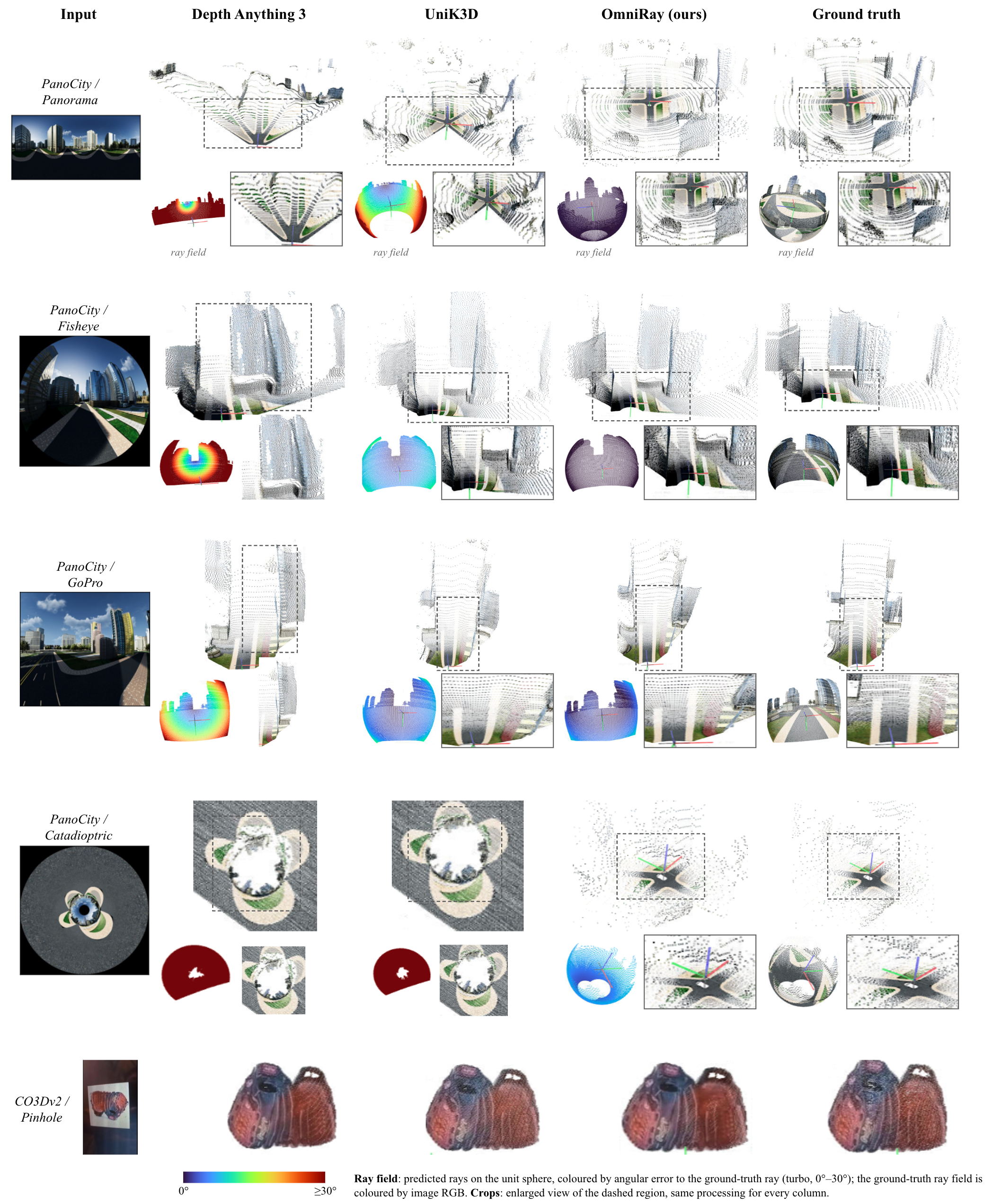}
\vspace{-15pt}
\caption{\textbf{Qualitative reconstructions on \underline{unseen} datasets.}
We compare OmniRay with Depth Anything 3 and UniK3D on PanoCity and CO3Dv2 across five camera configurations. Small visualizations show viewing-ray fields; larger renderings show the corresponding point clouds, with ground truth in the rightmost column. OmniRay better preserves scene structure in the panoramic and catadioptric examples without camera intrinsics or camera-type input.}
 \label{fig:qual}
\vspace{-10pt} 
\end{figure}

\begin{figure}[t]
    \centering
    \includegraphics[width=\linewidth]{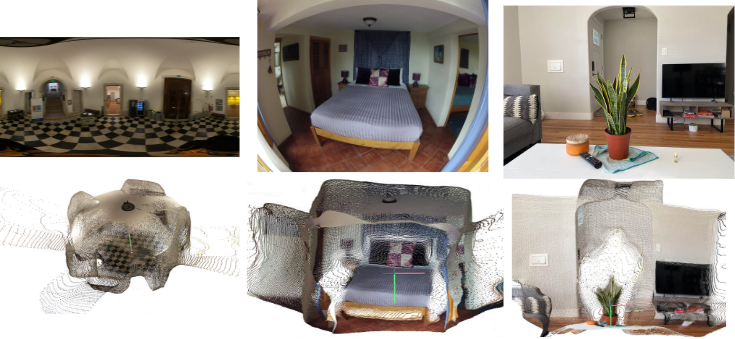}
\vspace{-20pt}
    \caption{\textbf{In-the-wild reconstructions.} Qualitative reconstructions on photographs collected from the internet, shown in order as a panorama, a fisheye image, and a pinhole image. Although these camera families are represented during training, the images come from different camera instances with unknown intrinsics and lens parameters. OmniRay requires no camera metadata at inference.}
    \vspace{-5pt}
    \label{fig:wild}
\end{figure}

Fig.~\ref{fig:qual} compares ray fields and camera-frame point clouds on unseen PanoCity and CO3Dv2 images, with each ray field colored by its angular error. On the panorama, OmniRay keeps the road continuous while UniK3D breaks it across the image and DA3 fails to recover the ground structure. On the fisheye and GoPro rows, OmniRay keeps the road, pavement, and lane markings straight where UniK3D bends them and DA3 also distorts the surrounding buildings. On the held-out catadioptric camera both baselines produce degraded reconstructions. On the CO3Dv2 pinhole example, OmniRay remains comparable to the baselines, demonstrating that the additional flexibility for wide-angle cameras does not come at the expense of pinhole imagery. On in-the-wild images, OmniRay also recovers coherent 3D structure, as shown in Fig. \ref{fig:wild}.

\paragraph{Ablation.}
Table~\ref{tab:abl} isolates and analyzes different ray representations.
The radial base performs well on fisheye and pinhole cameras ($1.761^\circ$ and $0.788^\circ$) but gives $21.825^\circ$ on panoramas, where a monotone radial profile assigns one incidence angle per image radius, an ordering that a panoramic projection does not follow (Fig.~\ref{fig:intuit}b). The spherical base reduces panoramic error to $5.088^\circ$, while the free-form field performs best among single representations on panoramas, fisheye, and pinhole, but reaches $11.335^\circ$ on the held-out catadioptric configuration. These results motivate combining a structured radial base with flexible residuals.

Adding an SH residual to the radial base improves panoramic and held-out accuracy; staged training achieves $0.610^\circ$ and $1.680^\circ$, respectively, and the best GoPro result ($3.393^\circ$). Adding the free-form residual further improves the panorama ($0.399^\circ$), fisheye ($1.322^\circ$), and catadioptric results ($1.538^\circ$) in the full staged model, which also outperforms both jointly trained three-term variants across all five configurations.
 
On pinhole cameras, the full model remains competitive at $0.418^\circ$. The additional spherical residual offers little benefit for this narrow-field geometry, while potentially providing greater flexibility across wide-field cameras. The VGGT-initialized variant~\citep{wang2025vggt} illustrates a similar trade-off: it achieves the best pinhole error ($0.330^\circ$) but performs worse than the full model on the other four configurations. Since VGGT is pretrained for multi-view geometry on perspective imagery, its initialization may favor narrow-field projections over the broader range of camera geometries considered here. As the backbone is fine-tuned end-to-end in both settings, this comparison suggests an effect of initialization rather than a limitation in model capacity.

Overall, these results support the staged radial $\rightarrow$ SH $\rightarrow$ free-form composition as a balance between geometric structure and representational flexibility.

\begin{table}[t!]
\centering
\caption{\textbf{Ray representation ablation.}
Mean angular ray error in degrees ($\downarrow$) at native resolution.
Seen camera families are evaluated on held-out configurations;
$\dagger$ denotes configurations excluded from training.
Only the Ray Module is trained and evaluated.
Best results are \textbf{bold}.}
\label{tab:abl}

\footnotesize
\setlength{\tabcolsep}{4pt}
\renewcommand{\arraystretch}{1.12}
\resizebox{0.75\linewidth}{!}{%
\begin{tabular}{@{}ll rrr rr@{}}
\toprule
& & \multicolumn{3}{c}{\textbf{Seen cameras}}
  & \multicolumn{2}{c}{\textbf{Held-out config}$^\dagger$} \\
\cmidrule(lr){3-5}
\cmidrule(l){6-7}
Ray representation & Training
& Panorama & Fisheye & Pinhole
& Catadioptric & GoPro \\
\midrule
Radial base
& --
& 21.825 & 1.761 & 0.788 & 6.039 & 5.377 \\

Spherical base
& --
& 5.088 & 1.540 & 0.790 & 13.327 & 5.962 \\

Free-form field
& --
& 0.783 & 1.508 & 0.621 & 11.335 & 5.191 \\

\addlinespace[3pt]
Radial + SH residual
& Joint
& 0.796 & 1.784 & 0.581 & 1.879 & 6.566 \\

Radial + SH residual
& Staged
& 0.610 & 1.545 & 0.675 & 1.680 & \textbf{3.393} \\

Radial + free-form
& Joint
& 1.018 & 1.394 & 0.403 & 3.038 & 3.667 \\

Radial + free-form
& Staged
& 1.111 & 1.605 & 0.386 & 1.797 & 4.248 \\

SH + free-form
& Joint
& 1.114 & 1.747 & 1.250 & 2.568 & 3.739 \\

SH + free-form
& Staged
& 1.026 & 2.567 & 0.553 & 2.316 & 5.142 \\

\addlinespace[3pt]
Radial + SH + free-form
& Joint
& 0.728 & 1.963 & 0.688 & 2.525 & 3.636 \\

Radial + SH + free-form
& Joint, $\gamma_0{=}\gamma_1{=}0$
& 0.451 & 1.864 & 0.705 & 1.838 & 3.604 \\

Radial + SH + free-form
& Staged, VGGT init
& 0.486 & 1.630 & \textbf{0.330} & 2.576 & 5.096 \\

\rowcolor{gray!10}
\textbf{Radial + SH + free-form}
& Staged
& \textbf{0.399} & \textbf{1.322} & 0.418
& \textbf{1.538} & 3.462 \\
\bottomrule
\end{tabular}%
}
\vspace{-10pt}
\end{table}

\section{Conclusion}
We began with two questions: how the pixel-to-ray mapping can remain both compact and expressive, and whether a single representation is necessary for every camera geometry. \textbf{OmniRay} explores both by composing the ray field. A monotone radial base captures the structure of conventional lenses, a spherical-harmonic residual models smooth deviations that a radial profile cannot express, and a free-form residual accounts for remaining local variation. Each component is applied as a rotation on $S^2$, allowing the representation to adapt to different camera geometries without camera-type input or model selection. A shared distance module scales the predicted rays into a point cloud in a single forward pass, and sampling lenses continuously during training helps the composition generalize to held-out camera configurations.

\begin{figure}[t]
    \centering
    \includegraphics[width=\linewidth]{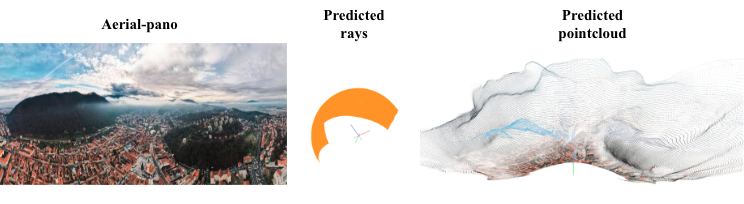}
\vspace{-20pt}
    \caption{Limitation: zero-shot in-the-wild reconstruction of an aerial panorama sourced from the internet. With the horizon near the top of the image, the predicted rays cover a bounded, forward-facing region.}
    \vspace{-10pt}
    \label{fig:limit}
\end{figure}

\textbf{Limitation.} A qualitative zero-shot example reveals a limitation on panoramic views that differ from the training distribution. The training images are captured near ground level, with horizons approximately aligned across images. As shown in Fig.~\ref{fig:limit}, on an in-the-wild aerial panorama with the horizon near the top of the frame, the predicted rays cover only a bounded, forward-facing region rather than the full viewing sphere. This example suggests that generalization depends not only on camera-projection diversity but also on the distribution of viewing altitudes represented during training.

\textbf{Future work.} A natural next step is to extend OmniRay to multi-view reconstruction and broaden the training distribution to include views with more varied camera orientations and altitudes.

\section*{Acknowledgments}
This research is based upon work supported by the Office of the Director of National Intelligence (ODNI), Intelligence Advanced Research Projects Activity (IARPA), via IARPA R\&D Contract No. 140D0423C0076. The views and conclusions contained herein are those of the authors and should not be interpreted as necessarily representing the official policies or endorsements, either expressed or implied, of the ODNI, IARPA, or the U.S. Government. The U.S. Government is authorized to reproduce and distribute reprints for Governmental purposes notwithstanding any copyright annotation thereon.

\subsection*{AI use statement}

In this work, we used generative AI tools to help with the implementation code and to polish grammar in the paper to improve readability. We have reviewed all AI-assisted work. We reviewed LLM-generated text to ensure it was consistent with our findings. We take responsibility for the final content of this work, including text, claims, or artifacts produced with the aid of generative AI.

\bibliography{iclr2027_conference}
\bibliographystyle{iclr2027_conference}

\newpage
\appendix
\section{Appendix}

\subsection{Implementation and training}
\label{sec:exp:impl}

\paragraph{Architecture.}
The encoder is DINOv2 ViT-L/14~\citep{oquab2023dinov2} with $1024$-dimensional features. The radial head is a two-layer multilayer perceptron (MLP) with width $256$ operating on the global token augmented with $(\log\text{aspect},\log\text{diagonal})$, while the spherical head is a one-layer MLP of the same width on the same input. The radial head predicts $K+8$ values with $K=16$ knots spanning $r\in[0,1.3]$ in half-diagonal units; the spherical head predicts $(L+1)^2\times3$ coefficients with $L=7$. These coefficients are shared across all pixels of an image. The free head is a convolutional decoder over the dense feature map. The Distance Module uses a width-$256$ DPT decoder~\citep{ranftl2021vision} over encoder blocks $\{4,11,17,23\}$ to predict the radial distance. Both residual heads are zero-initialized, while the radial head is initialized to a canonical camera, so training starts from a single radial lens.

\paragraph{Training.}

Images range from $896\times1792$ equirectangular panoramas to $512^2$ lens crops. We train with a token cap of $8192$ using a sampler for camera-type balancing. Within each image, we supervise 
$4096$ pixels drawn round-robin over the occupied cells of a $32\times32$ grid, ensuring coverage across the image. Training proceeds in two phases. We first train the Ray Module alone, activating its three components in order, and then attach the Distance Module at the final stage and train the full model end-to-end. In this second phase we use AdamW with weight decay $0.01$ and cosine decay after a warm-up, with learning rates of $5\!\times\!10^{-6}$ for the encoder and ray heads and $5\!\times\!10^{-5}$ for the distance decoder. The encoder runs in \texttt{bfloat16}, while the heads and losses remain in full precision. Training uses 8 NVIDIA A40 GPUs with distributed data parallelism.

\subsection{Experimental Details}
\label{sec:supp:details}

\paragraph{Oracle fitting protocol.} Each representation is fitted to the analytic ground-truth rays of a single camera by minimizing the mean cosine residual in Sec.~\ref{sec:method:loss}, with a multi-start sweep over the axis sign, roll and, where the representation uses one, the focal length. Every row is an independent fit under a fixed optimization budget, so each entry upper-bounds that representation's ceiling. The fitted error is the representation's approximation capacity for that camera and lower-bounds the error of any predictor built on it.

\begin{figure}[t!]
    \centering
    \includegraphics[width=\linewidth]{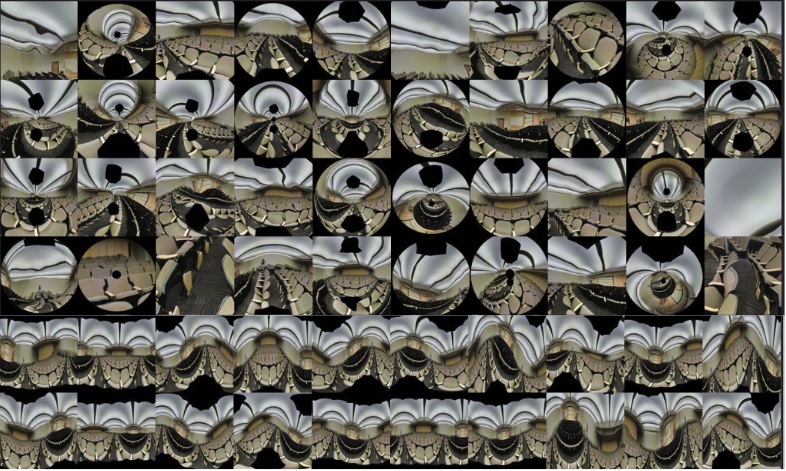}
\caption{\textbf{Random-lens augmentation.} Images rendered from a single panorama under cameras drawn from the sampling distribution of Sec.~\ref{sec:exp:data}. A camera is a viewing direction, a monotone radial profile and an imaged sensor region, so one family spans narrow perspective crops, wide-angle and fisheye views, off-center and anisotropic sensors, and annular domains. The projection varies continuously from sample to sample, so no fixed map from image position to ray direction fits more than one of them. Draws falling close to either held-out geometry are rejected and never enter training.}
\label{fig:randlens}
\end{figure}

\paragraph{Splits and Camera sampling.} For 2D3DS we train on areas 1-4 and evaluate on areas 5-6; for 360Loc we train on the Atrium and Piatrium scenes and evaluate on Concourse and Hall. Each synthetic camera samples a viewing orientation, an aspect ratio, a principal-point offset, an optional central blind spot, a rectangular or circular sensor boundary, and inner and outer incidence angles; panoramic cameras instead sample a vertical field of view, an azimuth and a small latitude stretch, with the horizon kept gravity-aligned as in the source captures. Draws near a held-out geometry are rejected: for the GoPro, any profile within $\delta=15^\circ$ of its incidence-angle curve at every radius; for the catadioptric camera, any sample with an inner angle in $[25^\circ,55^\circ]$ and a rim angle in $[140^\circ,170^\circ]$. Note: The Nayar catadioptric camera \citep{nayar1997catadioptric,baker1999theory} is a paraboloid mirror with orthographic projection, imaging a $40^\circ$--$155^\circ$ annulus about its axis; the mirror and camera share a single viewpoint. For PanoCity and CO3Dv2 we use the calibration provided with each dataset to obtain ground-truth rays.
Ground-truth rays and distances follow analytically from each sampled camera. Fig.~\ref{fig:randlens} shows images rendered under cameras drawn from this distribution.

\subsection{Additional Results}
\label{sec:supp:results}
\begin{figure}[t!]
    \centering
    \includegraphics[width=\linewidth]{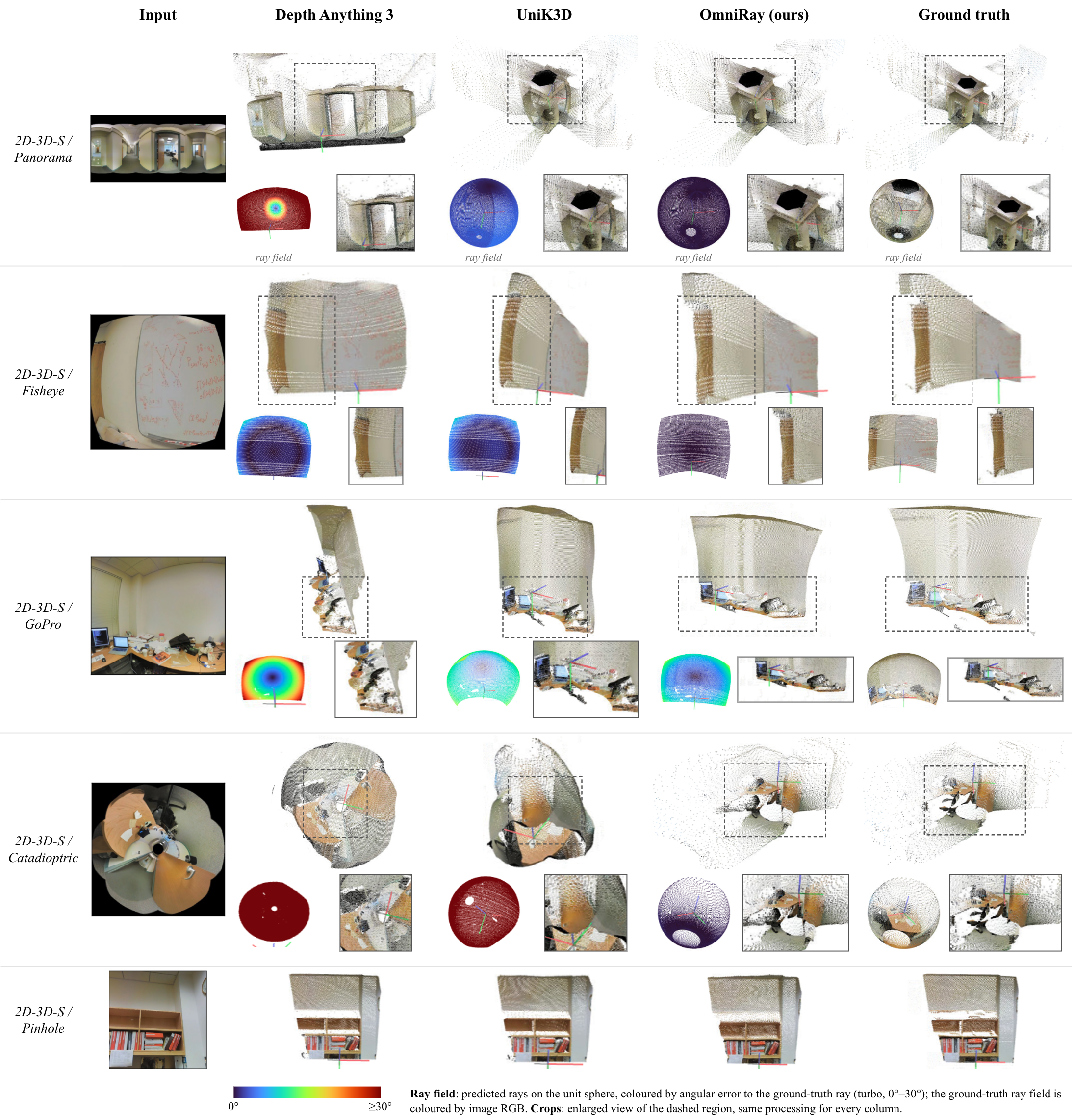}
\caption{\textbf{Qualitative reconstructions on \underline{seen} dataset: 2D3DS}}
 \label{fig:qual2d3ds}
\end{figure}

\begin{figure}[ht]
    \centering
    \includegraphics[width=\linewidth]{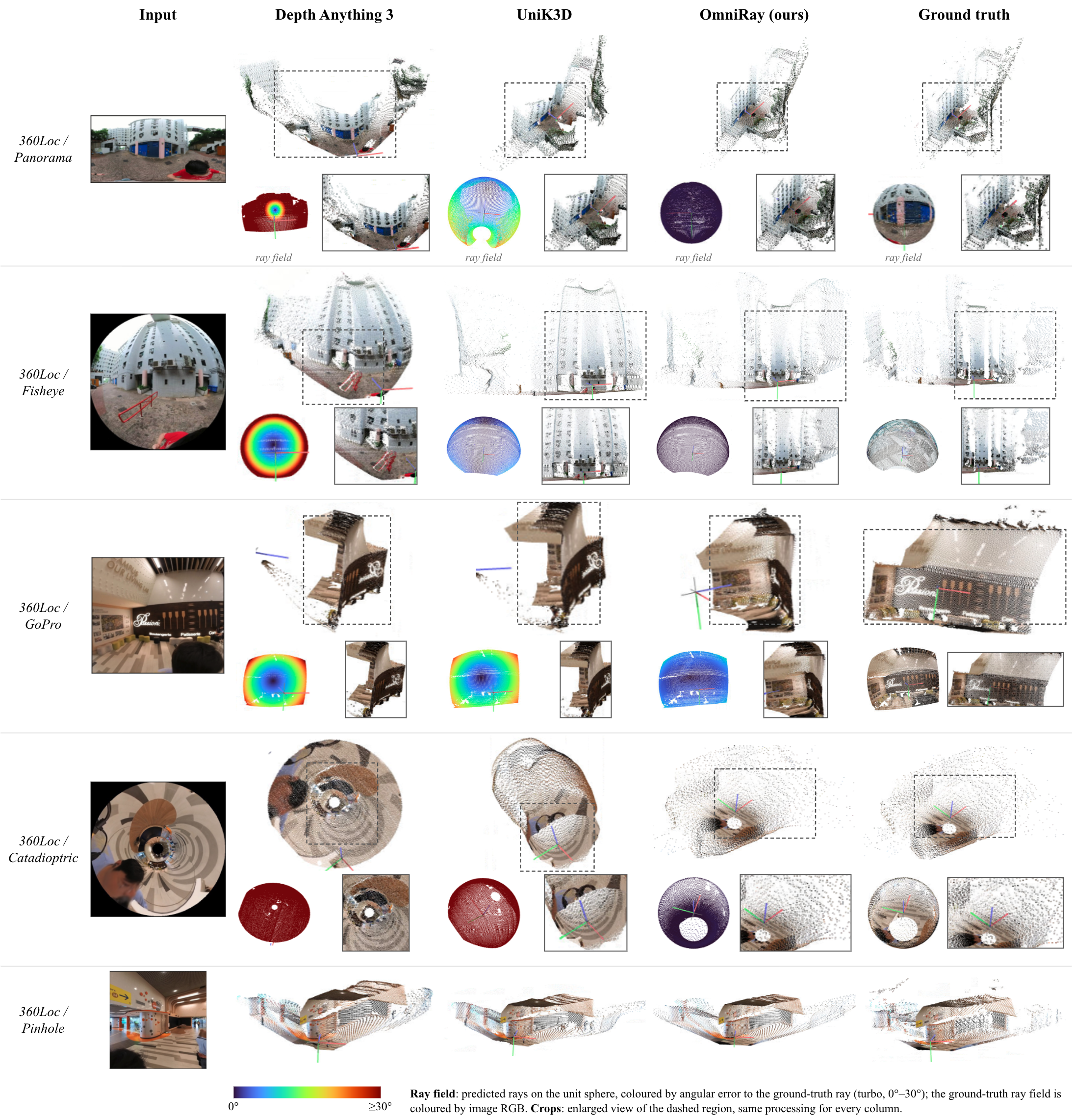}
\caption{\textbf{Qualitative reconstructions on \underline{seen} dataset: 360Loc}}
 \label{fig:qual360loc}
\end{figure}

\begin{figure}[ht]
    \centering
    \includegraphics[width=\linewidth]{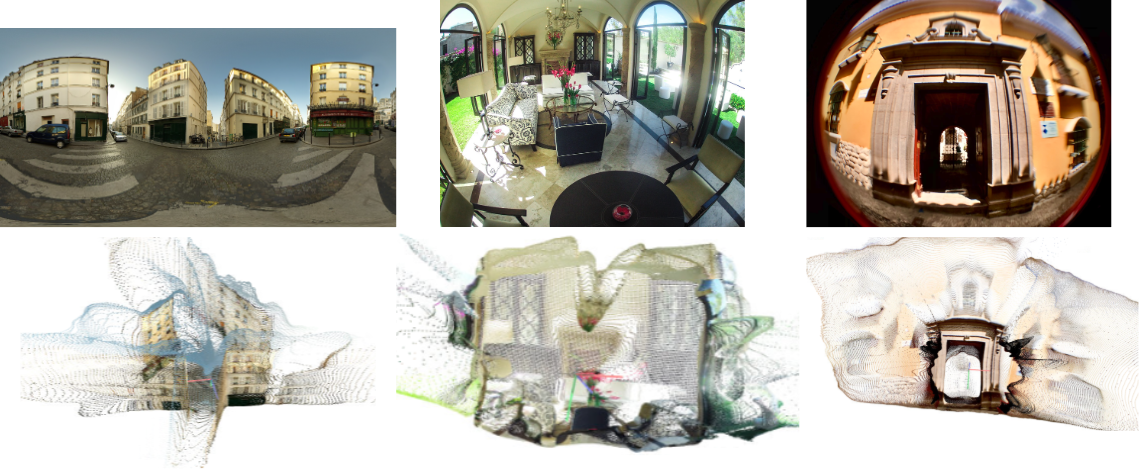}
    \caption{\textbf{Additional in-the-wild reconstructions.}}
    \label{fig:int_extra}
\end{figure}

Tables~\ref{tab:supp_cams}, \ref{tab:supp_scenes} and \ref{tab:supp_agg} report every metric of Sec.~\ref{sec:exp:bench} broken out per camera type, per dataset, and over the aggregate seen and held-out splits. Figures~\ref{fig:qual2d3ds} and \ref{fig:qual360loc} show reconstructions on the 2D3DS and 360Loc test scenes, which share the camera families used in training but none of their scenes, complementing the unseen-dataset comparison of Fig.~\ref{fig:qual}. Figure~\ref{fig:int_extra} provides additional visualizations of OmniRay on in-the-wild examples.

\begin{table}[t!]
\caption{Per-camera results, all metrics. Macro-averaged over test images. $^\dagger$ geometry
withheld from training. Baselines are median-scale-aligned per image. Acc, Comp and CD are medians of
the relative distance, divided by each scene's mean point norm. $\delta_1$ is $\delta<1.25$. F@10cm is the point-cloud F-score at a $10$\,cm distance threshold, the harmonic mean of precision and recall over nearest-neighbor distances. A dash means the method predicts no scene geometry.
}
\label{tab:supp_cams}
\begin{center}
\footnotesize
\setlength{\tabcolsep}{3.5pt}
\renewcommand{\arraystretch}{1.15}
\begin{tabular}{ll|c|cc|cccc}
\multicolumn{2}{c|}{} & \multicolumn{1}{c|}{\bf Ray} & \multicolumn{2}{c|}{\bf Radial Dist.} & \multicolumn{4}{c}{\bf point cloud Quality} \\
\multicolumn{1}{l}{\bf Camera} & \multicolumn{1}{l|}{\bf Method}
 & \multicolumn{1}{c|}{\bf ang$^\circ\downarrow$}
 & \multicolumn{1}{c}{\bf AbsRel$\downarrow$} & \multicolumn{1}{c|}{\bf $\delta_1\uparrow$}
 & \multicolumn{1}{c}{\bf CD$\downarrow$} & \multicolumn{1}{c}{\bf Acc$\downarrow$}
 & \multicolumn{1}{c}{\bf Comp$\downarrow$}
 & \multicolumn{1}{c}{\bf F@10cm$\uparrow$}
\\ \hline \\
pinhole & GeoCalib &   2.196 & -- & -- & -- & -- & -- & -- \\
        & AnyCalib &   1.043 & -- & -- & -- & -- & -- & -- \\
        & DA3      &   6.604 & 0.1508 & 0.862 & 0.0701 & 0.0862 & 0.0540 & 0.300 \\
        & UniK3D   &   3.619 & 0.1371 & 0.871 & 0.1058 & 0.1802 & 0.0315 & 0.356 \\
        & {\bf ours} & \textbf{0.155} & \textbf{0.0769} & \textbf{0.925} & \textbf{0.0242} & \textbf{0.0355} & \textbf{0.0130} & \textbf{0.533} \\
\\
fisheye & GeoCalib &  30.001 & -- & -- & -- & -- & -- & -- \\
        & AnyCalib &  22.134 & -- & -- & -- & -- & -- & -- \\
        & DA3      &  28.727 & 0.3225 & 0.562 & 0.1888 & 0.1111 & 0.2665 & 0.066 \\
        & UniK3D   &   7.055 & 0.2155 & 0.768 & 0.0543 & 0.0569 & 0.0517 & 0.198 \\
        & {\bf ours} & \textbf{0.154} & \textbf{0.0906} & \textbf{0.924} & \textbf{0.0159} & \textbf{0.0197} & \textbf{0.0121} & \textbf{0.531} \\
\\
panoramic & GeoCalib & 60.795 & -- & -- & -- & -- & -- & -- \\
        & AnyCalib &  62.788 & -- & -- & -- & -- & -- & -- \\
        & DA3      &  63.752 & 0.3352 & 0.547 & 0.3408 & 0.1956 & 0.4859 & 0.009 \\
        & UniK3D   &   8.336 & 0.2154 & 0.774 & 0.0678 & 0.0723 & 0.0633 & 0.174 \\
        & {\bf ours} & \textbf{0.110} & \textbf{0.0593} & \textbf{0.949} & \textbf{0.0137} & \textbf{0.0163} & \textbf{0.0111} & \textbf{0.585} \\
\\
catadioptric$^\dagger$
        & GeoCalib & 119.662 & -- & -- & -- & -- & -- & -- \\
        & AnyCalib & 107.129 & -- & -- & -- & -- & -- & -- \\
        & DA3      & 121.566 & 0.1617 & 0.771 & 1.0412 & 0.7232 & 1.3592 & 0.000 \\
        & UniK3D   &  75.702 & 0.2523 & 0.656 & 0.5948 & 0.5047 & 0.6849 & 0.004 \\
        & {\bf ours} & \textbf{1.120} & \textbf{0.0396} & \textbf{0.969} & \textbf{0.0131} & \textbf{0.0130} & \textbf{0.0131} & \textbf{0.633} \\
\\
GoPro$^\dagger$
        & GeoCalib &  14.126 & -- & -- & -- & -- & -- & -- \\
        & AnyCalib &  10.031 & -- & -- & -- & -- & -- & -- \\
        & DA3      &  16.637 & 0.1657 & 0.840 & 0.1097 & 0.0811 & 0.1383 & 0.079 \\
        & UniK3D   &   6.342 & 0.1665 & 0.857 & 0.0454 & 0.0449 & 0.0458 & \textbf{0.235} \\
        & {\bf ours} & \textbf{5.533} & \textbf{0.0842} & \textbf{0.932} & \textbf{0.0392} & \textbf{0.0354} & \textbf{0.0429} & 0.215 \\
\end{tabular}
\end{center}
\end{table}

\begin{table}[t!]
\caption{Per-dataset results, all metrics. Macro-averaged over test images. $^\dagger$ unseen in
training. Conventions as in Table \ref{tab:supp_cams}. CO3Dv2 contains perspective and fisheye
imagery only, so its zero-shot axis is the domain rather than the camera, and its object-masked
COLMAP ground truth covers roughly $12\%$ of pixels.}
\label{tab:supp_scenes}
\begin{center}
\footnotesize
\setlength{\tabcolsep}{3.5pt}
\renewcommand{\arraystretch}{1.15}
\begin{tabular}{ll|c|cc|cccc}
\multicolumn{2}{c|}{} & \multicolumn{1}{c|}{\bf Ray} & \multicolumn{2}{c|}{\bf Radial Dist.} & \multicolumn{4}{c}{\bf point cloud Quality} \\
\multicolumn{1}{l}{\bf Scenes} & \multicolumn{1}{l|}{\bf Method}
 & \multicolumn{1}{c|}{\bf ang$^\circ\downarrow$}
 & \multicolumn{1}{c}{\bf AbsRel$\downarrow$} & \multicolumn{1}{c|}{\bf $\delta_1\uparrow$}
 & \multicolumn{1}{c}{\bf CD$\downarrow$} & \multicolumn{1}{c}{\bf Acc$\downarrow$}
 & \multicolumn{1}{c}{\bf Comp$\downarrow$}
 & \multicolumn{1}{c}{\bf F@10cm$\uparrow$}
\\ \hline \\
2D3DS   & GeoCalib & 40.713 & -- & -- & -- & -- & -- & -- \\
        & AnyCalib & 37.520 & -- & -- & -- & -- & -- & -- \\
        & DA3      & 43.775 & 0.2124 & 0.747 & 0.3614 & 0.1744 & 0.5483 & 0.224 \\
        & UniK3D   & 19.432 & 0.1272 & 0.864 & 0.1487 & 0.0981 & 0.1994 & 0.411 \\
        & {\bf ours} & \textbf{1.619} & \textbf{0.0568} & \textbf{0.958} & \textbf{0.0236} & \textbf{0.0226} & \textbf{0.0246} & \textbf{0.767} \\
\\
360Loc  & GeoCalib & 48.223 & -- & -- & -- & -- & -- & -- \\
        & AnyCalib & 42.021 & -- & -- & -- & -- & -- & -- \\
        & DA3      & 48.064 & 0.2872 & 0.629 & 0.3212 & 0.2121 & 0.4303 & 0.048 \\
        & UniK3D   & 19.979 & 0.2709 & 0.676 & 0.1601 & 0.1455 & 0.1747 & 0.154 \\
        & {\bf ours} & \textbf{1.274} & \textbf{0.0713} & \textbf{0.947} & \textbf{0.0160} & \textbf{0.0160} & \textbf{0.0160} & \textbf{0.590} \\
\\
PanoCity$^\dagger$
        & GeoCalib & 47.132 & -- & -- & -- & -- & -- & -- \\
        & AnyCalib & 42.335 & -- & -- & -- & -- & -- & -- \\
        & DA3      & 50.533 & 0.1819 & 0.774 & 0.3677 & 0.3318 & 0.4037 & 0.001 \\
        & UniK3D   & 21.220 & 0.1940 & 0.815 & 0.2120 & 0.2719 & 0.1522 & 0.015 \\
        & {\bf ours} & \textbf{1.350} & \textbf{0.0823} & \textbf{0.914} & \textbf{0.0240} & \textbf{0.0333} & \textbf{0.0147} & \textbf{0.141} \\
\\
CO3Dv2$^\dagger$
        & GeoCalib &  3.713 & -- & -- & -- & -- & -- & -- \\
        & AnyCalib &  2.711 & -- & -- & -- & -- & -- & -- \\
        & DA3      & \textbf{2.041} & 0.0573 & 0.950 & 0.0288 & \textbf{0.0199} & 0.0377 & 0.282 \\
        & UniK3D   &  2.676 & \textbf{0.0478} & \textbf{0.963} & \textbf{0.0277} & 0.0233 & \textbf{0.0322} & \textbf{0.293} \\
        & {\bf ours} &  2.967 & 0.0889 & 0.914 & 0.0380 & 0.0291 & 0.0469 & 0.173 \\
\end{tabular}
\end{center}
\end{table}

\begin{table}[t!]
\caption{Aggregate results over the $4500$-sample benchmark. Conventions as in
Table \ref{tab:supp_cams}. Seen cameras are perspective, fisheye and panoramic; the held-out
geometries are the catadioptric mirror and the GoPro fisheye.}
\label{tab:supp_agg}
\begin{center}
\footnotesize
\setlength{\tabcolsep}{3.5pt}
\renewcommand{\arraystretch}{1.15}
\begin{tabular}{ll|c|cc|cccc}
\multicolumn{2}{c|}{} & \multicolumn{1}{c|}{\bf Ray} & \multicolumn{2}{c|}{\bf Radial Dist.} & \multicolumn{4}{c}{\bf point cloud Quality} \\
\multicolumn{1}{l}{\bf Split} & \multicolumn{1}{l|}{\bf Method}
 & \multicolumn{1}{c|}{\bf ang$^\circ\downarrow$}
 & \multicolumn{1}{c}{\bf AbsRel$\downarrow$} & \multicolumn{1}{c|}{\bf $\delta_1\uparrow$}
 & \multicolumn{1}{c}{\bf CD$\downarrow$} & \multicolumn{1}{c}{\bf Acc$\downarrow$}
 & \multicolumn{1}{c}{\bf Comp$\downarrow$}
 & \multicolumn{1}{c}{\bf F@10cm$\uparrow$}
\\ \hline \\
all     & GeoCalib & 45.356 & -- & -- & -- & -- & -- & -- \\
        & AnyCalib & 40.625 & -- & -- & -- & -- & -- & -- \\
        & DA3      & 47.457 & 0.2272 & 0.716 & 0.3501 & 0.2394 & 0.4608 & 0.091 \\
        & UniK3D   & 20.211 & 0.1974 & 0.785 & 0.1736 & 0.1718 & 0.1754 & 0.193 \\
        & {\bf ours} & \textbf{1.414} & \textbf{0.0701} & \textbf{0.940} & \textbf{0.0212} & \textbf{0.0240} & \textbf{0.0184} & \textbf{0.499} \\
\\
seen    & GeoCalib & 30.997 & -- & -- & -- & -- & -- & -- \\
        & AnyCalib & 28.655 & -- & -- & -- & -- & -- & -- \\
        & DA3      & 33.028 & 0.2695 & 0.657 & 0.1999 & 0.1310 & 0.2688 & 0.125 \\
        & UniK3D   &  6.336 & 0.1893 & 0.804 & 0.0760 & 0.1031 & 0.0488 & 0.242 \\
        & {\bf ours} & \textbf{0.140} & \textbf{0.0756} & \textbf{0.933} & \textbf{0.0179} & \textbf{0.0238} & \textbf{0.0120} & \textbf{0.550} \\
\\
held-out$^\dagger$
        & GeoCalib & 66.894 & -- & -- & -- & -- & -- & -- \\
        & AnyCalib & 58.580 & -- & -- & -- & -- & -- & -- \\
        & DA3      & 69.102 & 0.1637 & 0.806 & 0.5754 & 0.4021 & 0.7488 & 0.040 \\
        & UniK3D   & 41.022 & 0.2094 & 0.756 & 0.3201 & 0.2748 & 0.3654 & 0.120 \\
        & {\bf ours} & \textbf{3.326} & \textbf{0.0619} & \textbf{0.951} & \textbf{0.0261} & \textbf{0.0242} & \textbf{0.0280} & \textbf{0.424} \\
\end{tabular}
\end{center}
\end{table}

\end{document}